\newcommand{\CLASSINPUTinnersidemargin}{18mm}
\newcommand{\CLASSINPUToutersidemargin}{12mm}
\newcommand{\CLASSINPUTtoptextmargin}{20mm}
\newcommand{\CLASSINPUTbottomtextmargin}{25mm}
\documentclass[10pt,conference,a4paper]{IEEEtran}

\usepackage[utf8]{inputenc}

\usepackage{amsmath}

\usepackage{times}

\usepackage{graphicx}
\usepackage{subfigure}

\usepackage{multirow}

\DeclareGraphicsExtensions{.png,.eps,.ps,.pdf}

\usepackage{url}

\usepackage[english]{babel}

\begin{document}

\title{Classifying Suspicious Content in Tor Darknet}

\author{
\IEEEauthorblockN{Eduardo Fidalgo}
\IEEEauthorblockA{
Dept. IESA.\\
Universidad de León\\
Researcher at INCIBE\\
eduardo.fidalgo@unileon.es}

\and

\IEEEauthorblockN{Roberto A. Vasco-Carofilis}
\IEEEauthorblockA{
Dept. IESA.\\
Universidad de León\\
Researcher at INCIBE\\
rvasc@unileon.es}

\and

\IEEEauthorblockN{Francisco Jañez-Martino}
\IEEEauthorblockA{
Dept. IESA.\\
Universidad de León\\
Researcher at INCIBE\\
fjanm@unileon.es}

\and

\IEEEauthorblockN{Pablo Blanco-Medina}
\IEEEauthorblockA{
Dept. IESA.\\
Universidad de León\\
Researcher at INCIBE\\
pblanm@unileon.es}}

\maketitle

\begin{abstract}

One of the tasks of law enforcement agencies is to find evidence of criminal activity in the Darknet. However, visiting thousands of domains to locate visual information containing illegal acts manually requires a considerable amount of time and resources. Furthermore, the background of the images can pose a challenge when performing classification.

To solve this problem, in this paper, we explore the automatic classification Tor Darknet images using Semantic Attention Keypoint Filtering, a strategy that filters non-significant features at a pixel level that do not belong to the object of interest, by combining saliency maps with Bag of Visual Words (BoVW). We evaluated SAKF on a custom Tor image dataset against CNN features: MobileNet\_v1 and Resnet50, and BoVW using dense SIFT descriptors, achieving a result of $87.98$\% accuracy and outperforming all other approaches.
\end{abstract}


{\bf Type of contribution:}  {\it Research already published}

\section{Introduction}

The Darknet is the portion of the Deep Web that cannot be indexed by standard search engines, requiring unique browsers to access. The most famous Darknet is the Tor network, which provides its users with a layer of anonymity. However, this feature also enables criminal activity inside the Tor network, such as selling weapons and drugs, counterfeiting identity, among others~\cite{CREIC_CISIS:2017}.

The automatic recognition and classification of this content can be a time and resource demanding task for  Law Enforcement Agencies (LEA).

To support this task, in this paper~\cite{fidalgo2019classifying}, we introduced Semantic Attention Keypoint Filtering (SAKF), a method that improves the classification of Tor domain images into $5$ illicit categories by combining saliency maps with Bag of Visual Words (BoVW)~\cite{Csurka2004}.  Using this approach, the content of a Tor Hidden Service can be determined based on the its hosted images.


\section{Bag of Visual Words (BoVW)} \label{sec:bovm}

In the Bag of Visual Words (BoVW) model~\cite{Csurka2004}, each image is represented by the frequency of appearance of particular visual elements, called visual words.

First, a set keypoints are sampled from the image. Then, around each keypoint, a small squared region is considered and described through the use of a feature vector or descriptor. Finally, the BoVW computes a dictionary, i.e. a set of the visual words that could be present in a dataset. 

These visual words are the resulting clusters from a clustering process over the feature vectors extracted from the images belonging to the training set.

The keypoint sampling method and type of descriptors extracted from each patch may vary. In this work, we used a dense grid for sampling the key points, and SIFT (Scale-Invariant Feature Transform) descriptors \cite{Lowe2004}. To create the dictionary, we used K-means clustering.

During the stage of image representation, BoVW globally describes the images using the dictionary. First, a set of keypoints from each image is sampled and described with a feature vector. Then, these vectors are assigned to the closest visual word in terms of vector distance. Finally, the histogram of the visual words that represents all the initial feature vectors is used as the final image descriptor.

\section{Semantic Attention Key point Filtering} \label{sec:safk}

Previous work  \cite{Fidalgo2018_PRL} demonstrated that the accuracy in an image classification process, where BoVW and the saliency map are combined, depends on the standard deviation $\sigma$ of the Gaussian kernel used in the \textit{image signature} algorithm by Hou et al.~\cite{Hou}, also known as blurring factor. 

The saliency map algorithm proposed by Hou et al.~\cite{Hou} considers the image as the sum of the foreground and background signals. The foreground of an image can be computed as the sign map of the image’s DCT (discrete cosine transform) coefficients. 

Let $SM_{\sigma}$ be the image signature computed for a given blurring factor $\sigma$. We binarized  $SM_{\sigma}$ using the Otsu threshold~\cite{otsu1979threshold},  separating pixels into two classes, foreground and background. The result is  an image, $SM_{bin}$, where the attention zone corresponds to white pixels, and the black ones represent the background information.

Dense SIFT descriptors are extracted from the original image and separated into two groups: foreground, $d_F$, and background descriptors, $d_B$.  Subsequently, two different visual dictionaries are calculated using $d_F$ and $d_B$, called $VD_F$ and $VD_B$, respectively.

\begin{figure}[!htp]
\begin{center}
\includegraphics[width=1\linewidth]{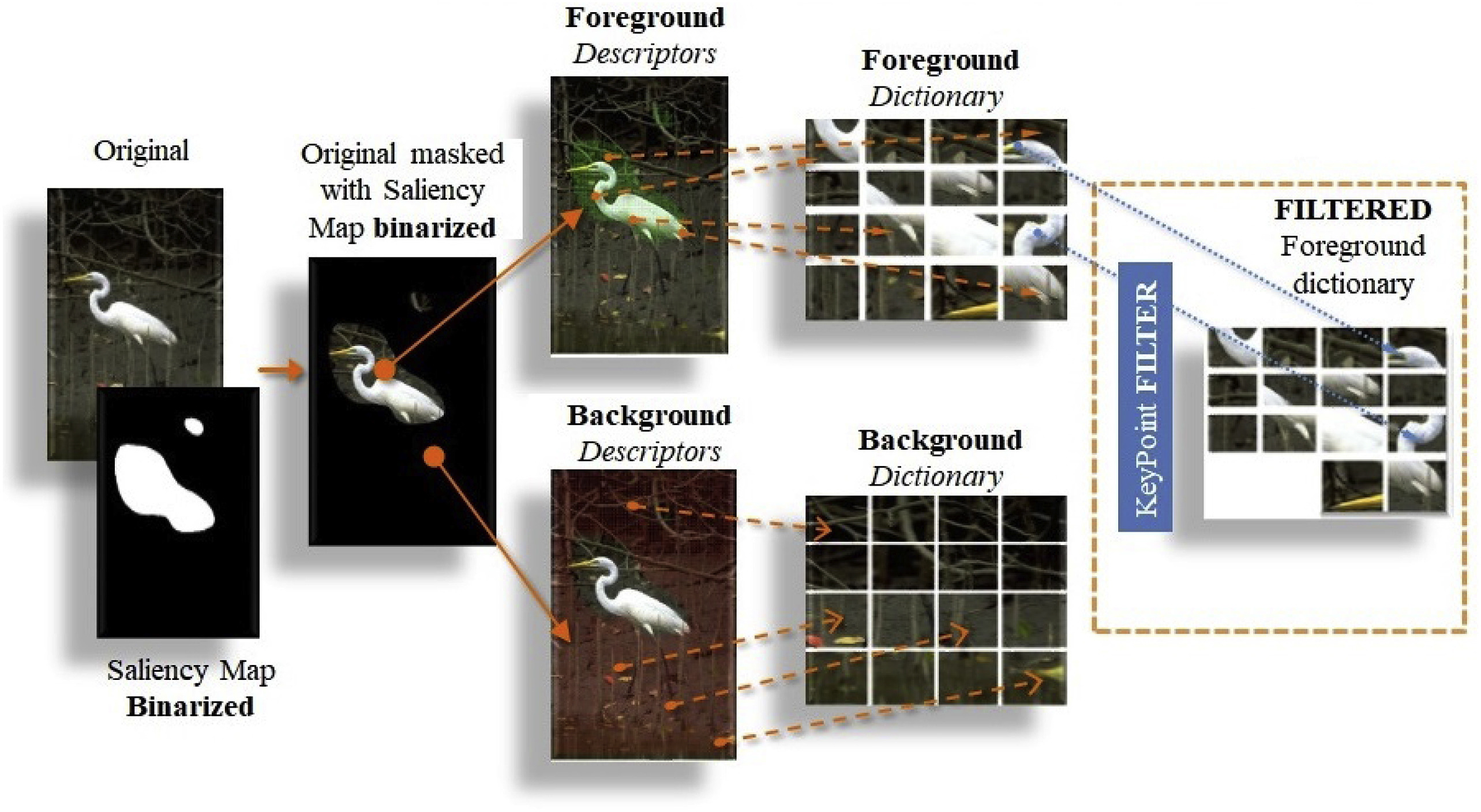}
\end{center}
\caption{Overview of the proposed SAKF method. 
}
\label{fig:SAKF_Visual_process}
\end{figure}

At this point, SAKF is applied to select foreground descriptors whose semantic meaning is closer to the main attention zone, i.e. the areas with values $1$ in $SM_{bin}$. For this, given an image, we make a semantic attention selection for all the foreground descriptors, and filter them using their Euclidean distances to the foreground and background dictionaries. Then, we generate the BoVW descriptor matrix only with the foreground descriptors whose measured distance is closer to the foreground dictionary. Fig.~\ref{fig:SAKF_Visual_process} visually depicts the SAKF method. 

\section{Experimental results} \label{sec:results}

To demonstrate the effectiveness of the proposed method, we conducted independent experiments using $7$ publicly available datasets, selecting $75\%$ of the data for the training process and $25\%$ of the data for testing. We included the TOr Images Categories (TOIC)~\cite{CREIC_CISIS:2017}, a dataset with $698$ real images taken from Tor domains, to apply SAKF to the task of Tor Darknet domain classification. This dataset contains $5$ categories of illegal content. The remaining $6$ datasets contain general purpose images. 

\begin{figure}[!htp]
\begin{center}
\includegraphics[width=0.8\linewidth]{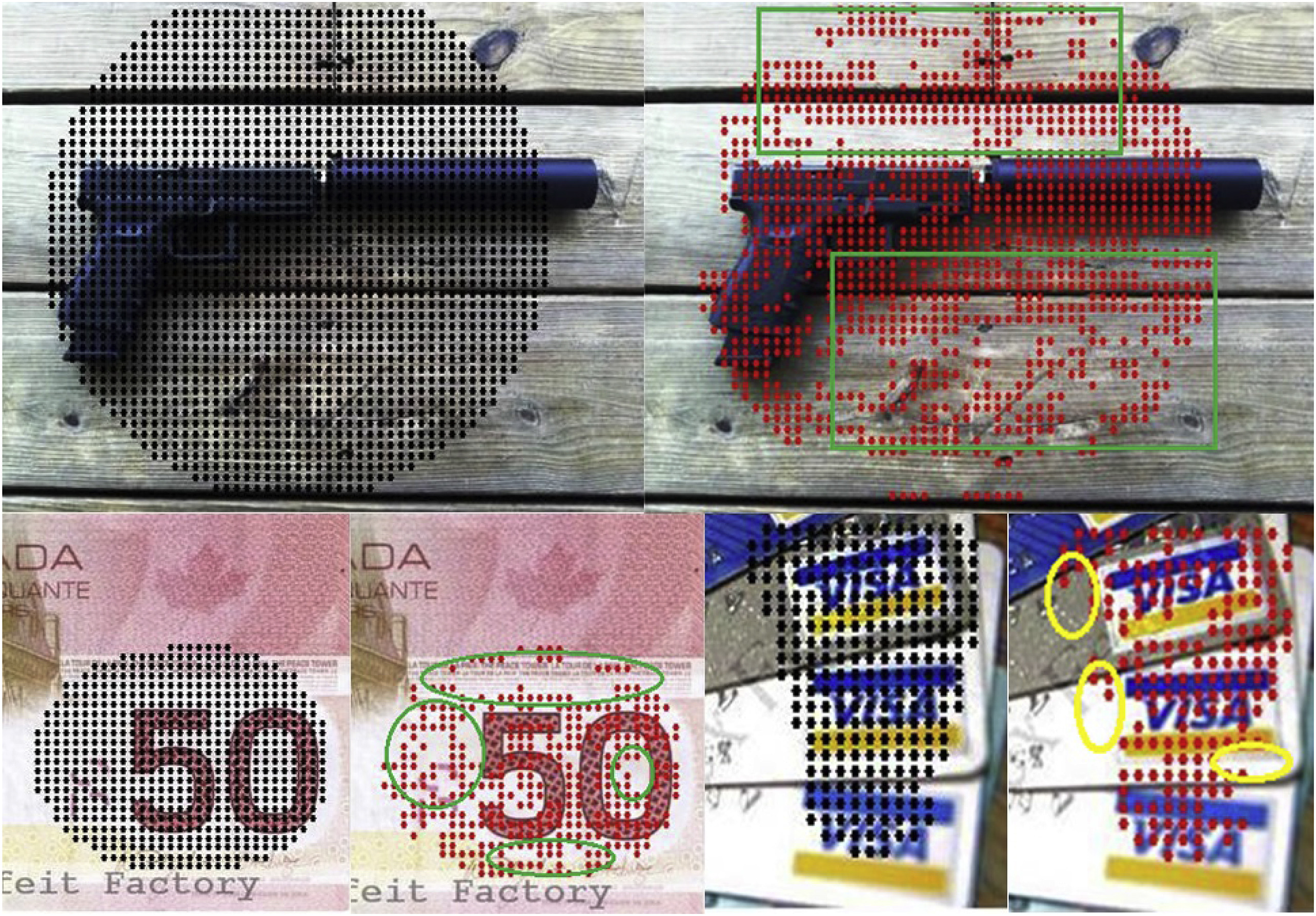}
\end{center}
\caption{Three samples from TOIC. Foreground descriptors are represented by black dots (left), while the red dots (right) represent SAKF descriptors.}
\label{fig:SAKF_filter}
\end{figure}

In Fig.~\ref{fig:SAKF_filter} we can see how SAKF filters out non-relevant information, in three examples taken from the TOIC dataset.

The system used for classification maintained the configuration of previous works~\cite{ CREIC_CISIS:2017,Fidalgo2018_PRL, Fidalgo2016, FidalgoRIAI2019}, i.e., dense SIFT descriptors~\cite{Lowe2004} with step and size $7$, K-means to obtain a $2048$ visual words dictionary, BoVW feature vectors~\cite{Csurka2004} built through a hard assignment approach and Support Vector Machine (SVM) classifier with a lineal kernel.

Finally, we made a comparison between the baseline from previous works~\cite{CREIC_CISIS:2017, Fidalgo2016, FidalgoRIAI2019, RubelBiswasEduardoFidalgo2017}, where we used a classical BoVW model together with dense SIFT, our proposal dSIFT+BoVW (SAKF), and Convolutional Neural Networks (CNN) features, extracting the last layer of two pretrained CNNs on ImageNet dataset: MobileNet\_v1 \cite{mobilenetv1} and ResNet50 \cite{resnet50}. The extracted features were used to train an SVM classifier with a linear kernel with five different sets of training and test, replicating the same conditions than when we used dense SIFT together with BoVW encoding.

The results of our comparisons are presented in Table~\ref{tab:comparision}.

\begin{table}[!htb]
\caption{Classification accuracy on TOIC dataset (\% of examples correctly classified).}
\begin{center}
\begin{tabular}{c c}
\hline
\hline
\textbf{Methods} & \textbf{Accuracy}\\
\hline
MobileNet\_v1 & $74.06$\%  \\
ResNet50 & $81.37$\%  \\
dSIFT + BoVW~\cite{CREIC_CISIS:2017} & $85.78$\%  \\
dSIFT + BoVW + SAKF \cite{fidalgo2019classifying} & $\mathbf{87.98}$\textbf{\%}  \\
\hline
\hline
\end{tabular}
\end{center}
\label{tab:comparision}
\end{table}

\section{Conclusions} \label{sec:conclusions}

In this paper, we presented Semantic Attention Keypoint Filtering (SAKF), a method that improves the image classification performance based on the Bag of Visual Words (BoVW) framework. Our proposal filters background information and extracts descriptors only from the foreground.

We compared the results of our method against similar implementations and CNN feature-based approaches, achieving higher accuracy than previous works.

\section*{Acknowledgements}

This work was supported by the framework agreement between the Universidad de Le\'{o}n and INCIBE (Spanish National Cybersecurity Institute) under Addendum $01$.
 
We acknowledge NVIDIA Corporation with the donation of the TITAN Xp and Tesla K40 GPUs used for this research.

\bibliographystyle{IEEEtran}
\bibliography{biblio}

\end{document}